# RePAD2: Real-Time, Lightweight, and Adaptive Anomaly Detection for Open-Ended Time Series

Ming-Chang Lee[1] and Jia-Chun Lin[2]

[1]Department of Computer science, Electrical engineering and Mathematical sciences, Høgskulen på Vestlandet (HVL), Bergen, Norway

[2]Department of Information Security and Communication Technology, Norwegian University of Science and Technology, Gjøvik, Norway

[1] ming-chang.lee@hvl.no
[2]jia-chun.lin@ntnu.no

# RePAD2: Real-Time, Lightweight, and Adaptive Anomaly Detection for Open-Ended Time Series

Ming-Chang Lee[1] [a] and Jia-Chun Lin[2] [b]

[1]*Department of Computer science, Electrical engineering and Mathematical sciences, Høgskulen på Vestlandet (HVL), Bergen, Norway*
[2]*Department of Information Security and Communication Technology, Norwegian University of Science and Technology (NTNU), Gjøvik, Norway*
mingchang1109@gmail.com, jia-chun.lin@ntnu.no

Abstract: An open-ended time series refers to a series of data points indexed in time order without an end. Such a time series can be found everywhere due to the prevalence of Internet of Things. Providing lightweight and real-time anomaly detection for open-ended time series is highly desirable to industry and organizations since it allows immediate response and avoids potential financial loss. In the last few years, several real-time time series anomaly detection approaches have been introduced. However, they might exhaust system resources when they are applied to open-ended time series for a long time. To address this issue, in this paper we propose RePAD2, a lightweight real-time anomaly detection approach for open-ended time series by improving its predecessor RePAD, which is one of the state-of-the-art anomaly detection approaches. We conducted a series of experiments to compare RePAD2 with RePAD and another similar detection approach based on real-world time series datasets, and demonstrated that RePAD2 can address the mentioned resource exhaustion issue while offering comparable detection accuracy and slightly less time consumption.

## 1 INTRODUCTION

A time series refers to a series of data points or observations obtained through repeated measurements over time (Ahmed et al. 2016). In the real world, many time series are continuously observed and collected. They are called open-ended time series in this paper because they do not have an end point. Such time series can be found everywhere due to the prevalence of the Internet of Things. Examples include CO2 levels measured by air quality monitors, human heart rates or blood pressures measured by medical IoT devices, electricity consumption by smart meters, humidity levels by smart agriculture IoT devices, water flow and saturation levels by smart ocean monitoring systems, etc.

Anomaly detection refers to a data analysis task that detects anomalous or abnormal data from a given dataset (Ahmed et al., 2016). An anomaly is defined as "an observation which deviates so much from other observations as to arouse suspicions that it was generated by a different mechanism" (Hawkins, 1980). Anomaly detection has been widely applied in many application domains such as intrusion detection, fraud detection, industrial damage, sensor network, healthcare, etc. (Hochenbaum, et al., 2017; Aggarwal and Yu, 2008; Xu and Shelton, ; Fisher et al., 2016; Wu et al., 2018; Staudemeyer, 2015; Bontemps et al., 2016). We believe that providing real-time and lightweight anomaly detection for open-ended time series is highly desirable to industry and organizations since it enables immediate response, allows appropriate countermeasure to be promptly taken, and avoids the occurrence of catastrophic failures/events.

During the past few years, several real-time and lightweight anomaly detection approaches have been introduced for time series, such as RePAD (Lee et al., 2020b) and its two successors ReRe (Lee et al., 2020a) and SALAD (Lee et al., 2021b). However, RePAD suffers from a resource exhaustion issue due to its design in determining its adaptive detection threshold. More specifically, the adaptive detection threshold will be periodically recalculated based on all previously derived average absolute relative error (AARE) values. Hence, when RePAD works on an open-ended time series for a long time, it will eventually exhaust the system resources. The same situation will also happen to ReRe and SALAD since they inherit the threshold design from RePAD.

---

[a]  https://orcid.org/0000-0003-2484-4366
[b]  https://orcid.org/0000-0003-3374-8536

To address the above-mentioned resource exhaustion issue, in this paper, we propose RePAD2 by improving RePAD in designing the adaptive detection threshold. Instead of relying on all historical AARE values to calculate the detection threshold, RePAD2 only calculates the detection threshold based on a fixed number of recently derived AARE values. In other words, we employ the concept of sliding window to redesign the threshold. However, it is unclear how different sliding window sizes impact the detection accuracy and time consumption of RePAD2.

To demonstrate the performance of RePAD2, we conducted a series of experiments based on real-world time series data from the Numenta Anomaly Benchmark (NAB, 2015) and compared RePAD2 with another two state-of-the-art real-time anomaly detection approaches. The experiment results show that RePAD2 with a sufficiently long sliding window can provide comparable prediction accuracy and slightly less time consumption.

The rest of the paper is organized as follows: Section 2 describes related work. Section 3 describes RePAD. In Section 4, we present how RePAD2 improves RePAD. Section 5 presents and discusses the experiments and the corresponding results. In Section 6, we conclude this paper and outline future work.

## 2 RELATED WORK

During the past decades, a number of anomaly detection approaches have been introduced, and they can be divided into two categories: Statistical learning approaches and machine-learning approaches.

Statistical learning approaches work by creating a statistical model for a set of normal data and then use the model to determine if a data point fits this model or not. If the data point has a low probability to be generated from the model, it is considered anomalous. For examples, AnomalyDetectionTs and AnomalyDetectionVec are statistical learning approaches proposed by (Twitter, 2015). AnomalyDetectionTs was designed to detect statistically significant anomalies in a given time series. On the other hand, AnomalyDetectionVec was designed to detect statistically significant anomalies in a given vector of observations without timestamp information. However, both approaches are parameter sensitive because they require human experts to set appropriate values to their parameters in order to achieve good detection performance.

Luminol (LinkedIn, 2018) is another statistical-based anomaly detection approach proposed by LinkedIn to identify anomalies in real user monitoring data. Given a time series, Luminol calculates an anomaly score for each data point. If a data point has a high score, it indicates that this data point is likely to be anomalous. However, human experts still need to further determine which data points are anomalies based on their domain knowledge and experiences. Siffer et al. (Siffer et al., 2017) introduced a time series anomaly detection approach based on Extreme Value Theory. This approach makes no assumption on the distribution of time series and requires no threshold manually set by humans. However, this approach needs a long period to do necessary calibration before conducting anomaly detection.

On the other hand, most machine learning-based anomaly detection approaches require either domain knowledge or human intervention. For instances, Yahoo proposed EGADS (Laptev et al., 2015) to detect anomalies on time series based on a collection of anomaly detection and forecasting models. However, EGADS requires to model the target time series so as to predict a data value later used by its anomaly detection module and its altering module. Lavin and Ahmad (Lavin and Ahmad, 2015) introduced Hierarchical Temporal Memory to capture pattern changes in time series. However, this approach requires 15% of its training dataset to be non-anomalous for training its neural network.

Greenhouse \(Lee et al., 2018) is an anomaly detection algorithm for time series based on Long Short-Term Memory (LSTM for short) (Hochreiter and Schmidhuber, 1997), which is an neural network designed to learn long short-term dependencies and model temporal sequences. Greenhouse requires all its training dataset to be non-anomalous. During the training phase, Greenhouse adopts a Look-Back and Predict-Forward strategy to detect anomalies. For a given time point, a window of most recently observed data point values of length $b$ is used as a Look-Back period to predict a subsequent window of data point values of length $f$. This feature enables Greenhouse to adapt to pattern changes in the training data. However, Greenhouse requires an offline training phase to train its LSTM model, and its detection threshold must be specified by human experts.

RePAD (Lee et al., 2020b) is one of the state-of-the-art real-time and lightweight time series anomaly detection approaches, and it is also based on LSTM and the Look-Back and Predict-Forward strategy. However, unlike Greenhouse, RePAD does not need an offline training phase. RePAD utilizes a simple LSTM network (with one hidden layer and ten hidden units) trained with short-term historic data points to predict upcoming data points, and then decides if each data point is anomalous based on a dynamic detection threshold that can adapt to pattern changes in the target time series. However, RePAD needs to recalculate its detection threshold based on all historical AARE values at every time point (except the first few time points). Hence, when RePAD works on an open-ended time series, it might eventually exhaust the underlying system resources.

ReRe (Lee et al., 2020a) is an enhanced real-time time series anomaly detection based on RePAD, and it was designed to keep a high true positive rate and a low false positive rate. It utilizes two LSTM models to jointly detect anomalous data points. One of the LSTM models works exactly as RePAD, whereas the other LSTM model works similar to RePAD but with a stricter detection threshold.

## 3 REPAD

Before introducing RePAD2, let us understand how RePAD works. Figure 1 illustrates the algorithm of RePAD (Lee et al., 2020b). RePAD uses short-term historical data points to predict upcoming data points by setting its Look-Back parameter (denoted by $b$) to a small integer and its Predict-Forward parameter (denoted by $f$) to 1. To help explain how RePAD works, let $b$ be 3. In other words, RePAD always predicts the next data point based on three historical data points.

When RePAD is launched, the current time point (denoted by $t$) is considered as 0. Since $b$ equals 3, RePAD needs to collect three data points to train an LSTM model. Hence, RePAD collects $v_0$, $v_1$, and $v_2$ at time points 0, 1, and 2, respectively. When $t$ is 2, RePAD can train the first LSTM model with data points $v_0$, $v_1$, and $v_2$. This model is denoted as $M$, and it is used by RePAD to predict the next data point, denoted by $\widehat{v_3}$. When $t$ advances to 3 and 4, RePAD continues the same process to predict $\widehat{v_4}$ and $\widehat{v_5}$, respectively (see lines 5 and 6 of Figure 1). When $t$ equals 5, RePAD can calculate $AARE_5$ based on Equation 1.

```
RePAD algorithm
Input: Data points in the target time series
Output: Anomaly notifications
Procedure:
1:   Let t be the current time point and t starts from 0; Let flag be True;
2:   While time has advanced {
3:      Collect data point v_t;
4:      if t ≥ b − 1 and t < 2b − 1 {  // i.e., 2 ≤ t < 5, if b = 3
5:         Train an LSTM model by taking [v_{t-b+1}, v_{t-b+2}, ..., v_t] as the training data;
6:         Let M be the resulting LSTM model and use M to predict v_{t+1};}
7:      else if t ≥ 2b − 1 and t < 2b + 1 {  // i.e., 5 ≤ t < 7, if b = 3
8:         Calculate AARE_t based on Equation 1;
9:         Train an LSTM model by taking [v_{t-b+1}, v_{t-b+2}, ..., v_t] as the training data;
10:        Let M be the resulting LSTM model and use M to predict v_{t+1};}
11:     else if t ≥ 2b + 1 and flag==True { //i.e., t ≥ 7 if b = 3
12:        if t ≠ 7 { Use M to predict v_t;}
13:        Calculate AARE_t based on Equation 1;
14:        Calculate thd based on Equation 2;
15:        if AARE_t ≤ thd{ v_t is not considered as an anomaly;}
16:        else{
17:           Train an LSTM model with [v_{t-b}, v_{t-b+1}, ..., v_{t-1}];
18:           Use the newly trained LSTM model to predict v_t;
19:           Calculate AARE_t using Equation 1;
20:           Calculate thd based on Equation 2;
21:           if AARE_t ≤ thd{ v_t is not considered as an anomaly;}
22:           else {
23:              v_t is reported as an anomaly immediately;
24:              Let flag be False;}}}
25:     else if t ≥ 2b + 1 and flag==False {
26:        Train an LSTM model with [v_{t-b}, v_{t-b+1}, ..., v_{t-1}];
27:        Use the newly trained LSTM model to predict v_t;
28:        Calculate AARE_t based on Equation 1;
29:        Calculate thd based on Equation 2;
30:        if AARE_t ≤ thd{
31:           v_t is not considered as an anomaly;
32:           Replace M with the new LSTM model from line 26;
33:           Let flag be True;}
34:        else {
35:           v_t is reported as an anomaly immediately; Let flag be False;}}}
```

Figure 1: The algorithm of RePAD (Lee et al., 2020b).

$$AARE_t = \frac{1}{b} \cdot \sum_{y=t-b+1}^{t} \frac{|v_y - \widehat{v_y}|}{v_y}, t \geq 2b-1 \quad (1)$$

where $v_y$ is the observed data value at time point $y$, and $\widehat{v_y}$ is the predicted data value for time point $y$. Recall $b$ equals 3, $AARE_5 = \frac{1}{3} \cdot \sum_{y=3}^{5} \frac{|v_y - \widehat{v_y}|}{v_y} = \frac{1}{3} \cdot \left( \frac{|v_3 - \widehat{v_3}|}{v_3} + \frac{|v_4 - \widehat{v_4}|}{v_4} + \frac{|v_5 - \widehat{v_5}|}{v_5} \right)$.

Note that RePAD always calculates $AARE_t$ based on three most recently absolute relative errors. A low AARE value indicates accurate prediction because the predicted values are close to the observed values. When $t$ advances to 6, RePAD keeps calculating $AARE_6$ and training a new LSTM model to predict $\widehat{v_7}$ (see lines 8 to 10 of Figure 1).

When $t$ advances to 7, RePAD calculates $AARE_7$ and then calculates its adaptive detection threshold, denoted by $thd$, (see line 14) using Equation 2 based on the Three-Sigma Rule (Hochenbaum et al., 2017). In other words, RePAD needs at least three AARE values to determine $thd$.

$$thd = \mu_{AARE} + 3 \cdot \sigma, t \geq 2b + 1 \quad (2)$$

where $\mu_{AARE}$ is the average AARE, and it is calculated as below.

$$\mu_{AARE} = \frac{1}{t-b-1} \cdot \sum_{x=2b-1}^{t} AARE_x \qquad (3)$$

In Equation 2, $\sigma$ is the standard deviation, and it can be derived as below.

$$\sigma = \sqrt{\frac{\sum_{x=2b-1}^{t}(AARE_x - \mu_{AARE})^2}{t-b-1}} \qquad (4)$$

Once the detection threshold $thd$ is calculated, it is used immediately by RePAD to decide if the upcoming data point is anomalous or not. If $AARE_t$ is not higher than the threshold (see line 15), $v_t$ is not considered as anomalous, and the current LSTM model will be kept for future prediction.

On the other hand, if $AARE_t$ is higher than the threshold (see line 16), it might indicate that either the data pattern of the time series has changed or that anomalies might have occurred. In this case, RePAD will try to adapt to the potential pattern change by retraining another new LSTM model to re-predict $\hat{v}_t$ and see if this newly trained LSTM model can lower $v_t$. If the new $AARE_t$ does not exceed the threshold (see line 21), RePAD does not consider $v_t$ anomalous. However, if the new $AARE_t$ exceeds the threshold, meaning that the newly trained LSTM model still cannot accurately predict the data point, RePAD immediately reports the data point as an anomaly, allowing for corresponding actions or countermeasures to be taken. Furthermore, RePAD sets its flag to false (see line 24), enabling a new LSTM model to be trained at the next time point instead of using the current LSTM model. The above process will repeat as time advanced.

## 4  REPAD2

As described previously, RePAD has several good features, including lightweight LSTM network (one hidden layer with 10 hidden units), dynamic detection threshold that can adapt to pattern changes, LSTM model training only in the beginning and only when AARE exceeds the detection threshold, etc. However, as shown in Equations 3 and 4, RePAD relies on all historical AARE values to calculate its detection threshold at every time point (except in the beginning phase between time point 0 and time point $2b$). This design might slow down RePAD when RePAD is employed to detect anomalies in an open-ended time series for a long time, and it might eventually exhaust the underlying system resources (e.g., the system memory) due to the increasing number of the historical AARE values.

```
Input: Data points in the target time series
Output: Anomaly notifications
Procedure:
1:   Let T be the current time point and T starts from 0; Let flag* be True;
2:   While time has advanced {
3:       Collect data point D_T;
4:       if T ≥ 2 and T < 5 {
5:           Train an LSTM model by taking D_{T-2}, D_{T-1}, and D_T as the training data;
6:           Let M* be the resulting LSTM model and use M* to predict D̂_{T+1};}
7:       else if T ≥ 5 and T < 7 {
8:           Calculate AARE*_T based on Equation 5;
9:           Train an LSTM model by taking D_{T-2}, D_{T-1}, and D_T as the training data;
10:          Let M* be the resulting LSTM model and use M* to predict D̂_{T+1};}
11:      else if T ≥ 7 and flag*==True {
12:          if T ≠ 7 { Use M* to predict D̂_T;}
13:          Calculate AARE*_T based on Equation 5;
14:          Calculate Thd* based on Equation 6;
15:          if AARE*_T ≤ Thd* { D_T is not considered as an anomaly;}
16:          else {
17:              Train an LSTM model with D_{T-3}, D_{T-2}, and D_{T-1};
18:              Use the newly trained LSTM model to re-predict D̂_T;
19:              Re-calculate AARE*_T using Equation 5;
20:              Re-calculate Thd* based on Equation 6;
21:              if AARE*_T ≤ Thd* {
22:                  D_T is not considered as an anomaly;
23:                  Replace M* with the new LSTM model from line 17;
24:                  Let flag* be True;}
25:              else {
26:                  D_T is reported as an anomaly immediately;
27:                  Let flag* be False;}}}
28:      else if T ≥ 7 and flag*==False {
29:          Train an LSTM model with D_{T-3}, D_{T-2}, and D_{T-1};
30:          Use the newly trained LSTM model to predict D̂_T;
31:          Calculate AARE*_T based on Equation 5;
32:          Calculate Thd* based on Equation 6;
33:          if AARE*_T ≤ Thd* {
34:              D_T is not considered as an anomaly;
35:              Replace M* with the new LSTM model from line 29;
36:              Let flag* be True;}
37:          else {
38:              D_T is reported as an anomaly immediately; Let flag* be False;}}}
```

Figure 2: The algorithm of RePAD2.

Figure 2 illustrates the algorithm of RePAD2 where $T$ denotes the current time point ($T$ starts from 0), $D_T$ denotes the observed data point at time point $T$, and $\widehat{D_T}$ denotes the predicted data point at time point $T$. It is clear that the algorithm of RePAD2 is similar to that of RePAD. However, instead of providing the flexibility to configure the Look-Back parameter, RePAD2 always uses three historical data points to predict each upcoming data point by following the Look-Back parameter suggestion made by (Lee et al. 2021a).

Similar to RePAD, RePAD2 always uses three historical absolute relative errors to calculate the average absolute relative error at time point $T$, denoted by $AARE_T^*$. The equation is shown as below.

$$AARE_T^* = \frac{1}{3} \cdot \sum_{y=T-2}^{T} \frac{|D_y - \widehat{D_y}|}{D_y}, T \geq 5 \qquad (5)$$

To address the resource exhaustion issue, RePAD2 uses Equation 6 to calculate its detection threshold at time point $T$ (where $T \geq 7$). The threshold is denoted by $Thd^*$.

$$Thd^* = \mu_{AARE}^* + 3 \cdot \sigma^*, T \geq 7 \qquad (6)$$

where $\mu^*_{AARE}$ and $\sigma^*$ are calculated via Equations (7) and (8), respectively.

$$\mu^*_{AARE} = \begin{cases} \frac{1}{T-4} \cdot \sum_{x=5}^{T} AARE^*_x, 7 \leq T < W+4 \\ \frac{1}{W} \cdot \sum_{x=T-W+1}^{T} AARE^*_x, T \geq W+4 \end{cases} \quad (7)$$

$$\sigma^* = \begin{cases} \sqrt{\frac{\sum_{x=5}^{T}(AARE^*_x - \mu^*_{AARE})^2}{T-4}}, 7 \leq T < W+4 \\ \sqrt{\frac{\sum_{x=T-W+1}^{T}(AARE^*_x - \mu^*_{AARE})^2}{W}}, T \geq W+4 \end{cases} \quad (8)$$

Note that, in Equations 7 and 8, $W$ is an integer to indicate how many historical AARE values will be considered to calculate $Thd^*$. If the total number of all historical AARE values is less than $W$, all the historical AARE values will be considered to calculate $Thd^*$. Otherwise, only the $W$ most recently derived AARE values will be used to calculate $Thd^*$. For instance, if $W$ equals 1000, then $Thd^*$ at time point 1004 will be calculated as $\frac{AARE^*_5 + AARE^*_6 + \cdots + AARE^*_{1004}}{1000} + 3 \cdot \sqrt{\frac{(AARE^*_5 - \mu^*_{AARE})^2 + \cdots + (AARE^*_{1004} - \mu^*_{AARE})^2}{1000}}$. Recall that $Thd^*$ will be re-calculated at every time point (except from time point 0 to time point 6). By restricting the number of AARE values to calculate $Thd^*$, we avoid RePAD2 from running out of the underlying system resources.

Another difference between RePAD2 and RePAD is that RePAD2 attempts to reduce unnecessary LSTM model training when a re-calculated $AARE^*$ value is not higher than $Thd^*$ (see lines 23 and 24 of Figure 2). In the next section, we will evaluate the performance of RePAD2 and investigate how different values of $W$ impact RePAD2.

## 5 EXPERIMENT RESULTS

To evaluate RePAD2, we compared it with another two state-of-the-art real-time and lightweight anomaly detection approaches (Lee et al.,2020b) and ReRe (Lee et al.,2020a) by conducting two experiments. In the first experiment, we chose one time series data called ec2-cpu-utilization-825cc2 (CC2 for short) from the Numenta Anomaly Benchmark (NAB, 2015). In the second experiment, we chose another time series data called rds-cpu-utilization-e47b3b (B3B for short) from the same benchmark. Both time series consist of 4032 data points that were collected every five minutes. However, CC2 was collected from April 10th to April 24th in 2014, whereas B3B was collected from April 10th to April 23rd in the same year. CC2 contains two point anomalies and one sequential anomaly noted by domain experts, whereas B3B contains one point anomaly and one sequential anomaly noted by domain experts. Note that a point anomaly is considered as a sequential anomaly of size one (Schneider et al., 2021).

Due to lack of open-source time series that is open-ended and contains anomalies indicated by domain experts, we created a long time series (called CC2-10) by duplicating CC2 ten times and concatenating the ten series together. We also did the same for B3B and created a ten-time longer time series called B3B-10. The purpose is to evaluate the performance of the three approaches (i.e., RePAD2, RePAD, and ReRe) on such long time series. Table 1 summaries the details of the two extended time series. There are 20 point anomalies and 10 sequential anomalies in CC2-10, and 10 point anomalies and 10 sequential anomalies in B3B-10. Figures 3 and 4 illustrate the two time series. Each point anomaly is marked as a red circle, and each sequential anomaly is marked in red.

Table 1: Details of the two extended time series used in the experiments.

| Name | CC2-10 | B3B-10 |
| --- | --- | --- |
| # of data points | 40320 | 40320 |
| Time interval (min) | 5 | 5 |
| # of point anomalies | 20 | 20 |
| # of sequential anomalies | 10 | 10 |

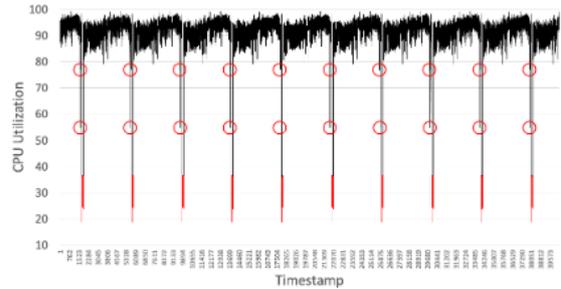

Figure 3: All data points of CC2-10. Note that all anomalies are marked in red.

To achieve a fair comparison, all the three approaches had the same hyperparameter and parameter setting as listed in Table 2, and they were implemented in DL4J (Deeplearning4j, 2023), which is a programming library written in Java for deep learning. All approaches adopted Early Stopping (EarlyStopping, 2023) to automatically determine the number of epochs (up to 50) for LSTM model training. Furthermore, the Look-Back parameter for both RePAD and ReRe was set to three so that all the three approaches always use three historical data points to predict each upcoming data point. In both experiments, RePAD2 was evaluated under four values for variable $W$: 1440, 4032, 8064, and 16128. These sliding window sizes are equivalent to 5, 14, 28, and 56 days because the number of data points collected per day in both CC2 and B3B was 288. All the experiments were performed on a laptop running MacOS Monterey 12.6 with 2.6 GHz 6-Core Intel Core i7 and 16GB DDR4 SDRAM.

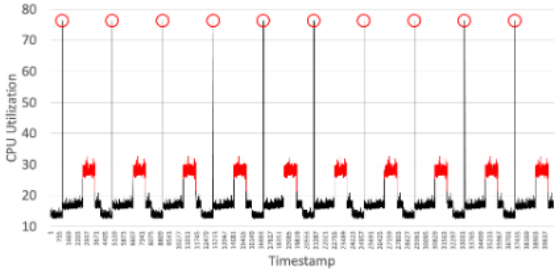

Figure 4: All data points of B3B-10. Note that all anomalies are marked in red.

Table 2: The hyperparameter and parameter setting used by the three approaches.

| Hyperparameters and parameters | Value |
| --- | --- |
| The number of hidden layers | 1 |
| The number of hidden units | 10 |
| The number of epochs | 50 |
| Learning rate | 0.005 |
| Activation function | tanh |
| Random seed | 140 |

To evaluate the detection accuracy of the three approaches, we followed the evaluation method used by (Lee et al., 2020a) to measure precision (which equals $TP/(TP+FP)$), recall (which equals $TP/(TP+FN)$), and F-score (which equals $2\times(Precision\times Recall)/(Precision+Recall)$). Note that $TP$, $FP$, and $FN$ represent true positive, false positive, and false negative, respectively. More specifically, if any anomaly occurring at time point $Z$ can be detected within a time period ranging from time point $Z-K$ to time point $Z+K$, we say that this anomaly is correctly detected. Note that we followed (Ren et al., 2019) and set $K$ to 7. This setting was applied to all the three approaches as well.

## 5.1 Experiment 1

Table 3 lists the detection accuracy of RePAD2, RePAD, and ReRe on CC2-10. When $W$ was 1440, RePAD2 has a poor precision (i.e., 0.531). Even though RePAD2 under this sliding window size has recall of 1, the low precision resulted in the worst F-score among all compared approaches. When $W$ was increased to 4032, the precision of RePAD2 significantly increased to 0.972, but its recall reduced to 0.7, leading to the F-score of 0.814. We can see that the detection accuracy of RePAD2 remained similar when $W$ was further increased to 8064, but it dropped slightly when $W$ was further increased to 16128. Based on the results shown in Table 3, we can see that RePAD2 provides a slightly better detection accuracy than RePAD when $W$ is 4032, 8064, or 16128. As compared with ReRe, RePAD2 offers a comparable detection accuracy when $W$ is 4032 or 8064.

To explain why RePAD2 has different results, Figures 5-8 illustrate how the detection threshold of RePAD2 changes over time under different values of $W$. When $W$ was 1440, we can see that the threshold curve as shown in Figure 5 apparently rises and then falls repeatedly, implying that RePAD2 periodically lost the memory about older historical AARE values and obtained the memory about newer historical AARE values. Due to the fact that the detection threshold is calculated at every time point based on the past 1440 AARE values, the threshold is prone to be affected by high AARE values. Such a phenomena caused a lot of false positives. That is why RePAD2 with $W$ of 1440 has the poor precision.

Table 3: The detection accuracy of different approaches on CC2-10.

| Approach | Precision | Recall | F-score |
| --- | --- | --- | --- |
| RePAD2 ($W$=1440) | 0.531 | 1 | 0.694 |
| RePAD2 ($W$=4032) | 0.972 | 0.7 | 0.814 |
| RePAD2 ($W$=8064) | 0.971 | 0.7 | 0.814 |
| RePAD2 ($W$=16128) | 0.965 | 0.7 | 0.811 |
| RePAD | 0.964 | 0.7 | 0.811 |
| ReRe | 0.971 | 0.7 | 0.814 |

When $W$ was increased to 4032 or 8064, the threshold curve became flatter (see Figures 6 and 7) since RePAD2 used more historical AARE values to calculate its detection threshold. In other words, the threshold is less affected by few high AARE values. However, when $W$ was further increased to 16128, it cannot further improve the precision of RePAD2 due to false positives. Nevertheless, it achieves a similar detection accuracy as RePAD. Therefore, according to the results, it is recommended to choose a sufficiently long sliding window (e.g., 4032) for RePAD2 so as to reduce false positives.

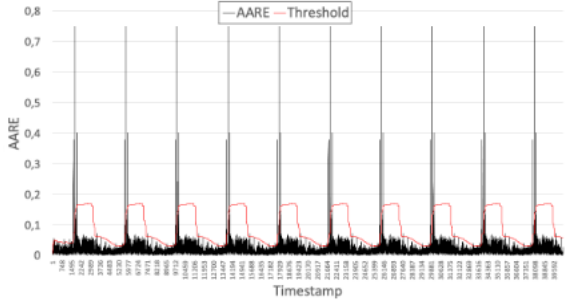

Figure 5: All derived AARE values vs. the detection threshold over time when RePAD2 worked on CC2-10 and had a sliding window of 1440.

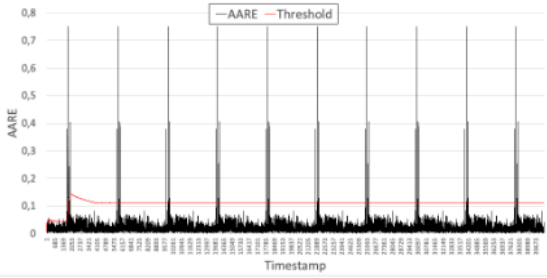

Figure 6: All derived AARE values vs. the detection threshold over time when RePAD2 worked on CC2-10 and had a sliding window of 4032.

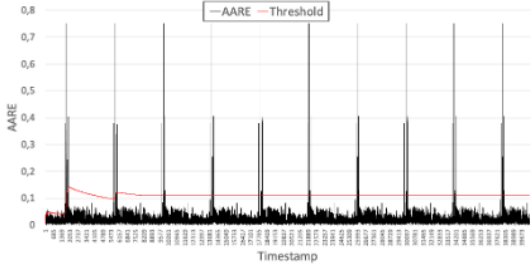

Figure 7: All derived AARE values vs. the detection threshold over time when RePAD2 worked on CC2-10 and had a sliding window of 8064.

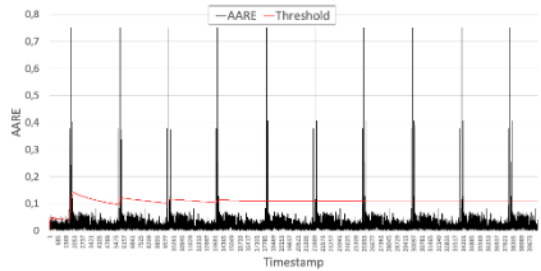

Figure 8: All derived AARE values vs. the detection threshold over time when RePAD2 worked on CC2-10 and had a sliding window of 16128.

Recall that RePAD2, RePAD, and ReRe are all designed to decide if each upcoming data point in the target time series is anomalous. When they find that their current LSTM models cannot accurately predict a data point, they will retrain a new LSTM model. Table 4 lists the LSTM retraining ratios of all the three approaches. When $W$ was 1440, RePAD2 required to retrain a LSTM model for 555 data points. Since the total number of the data points in CC2-10 is 40320, the LSTM model retraining ratio is 0.014. We can see that the retraining ratio of RePAD2 reduced when $W$ was increased, implying that including more historical AARE values to calculate the threshold helps reduce LSTM model retraining. When RePAD2 is compared with RePAD, it requires slightly more model retraining. But when it is compared with ReRe, RePAD2 has a lower retraining ratio. This is because ReRe employs two detectors to jointly detect anomalies. The detection threshold used by detector 2 is more stricter than the detection threshold used by detector 1, which caused more LSTM model retraining.

Table 5 shows the average time required by the three approaches to decide if a data point in CC2-10 is anomalous while LSTM model retraining is required. We can see that the time required by RePAD2 slightly reduced when $W$ was increased, implying that including more AARE values to calculate the detection threshold can slightly help reduce the time consumption of RePAD2. As compared with RePAD, RePAD2 is slightly more efficient when $W$ is 8064 or 16128. On the other hand, ReRe consumes more time than RePAD2 and RePAD since it employs two parallel detectors (rather than one detector) to detect anomalies simultaneously.

Table 6 lists the time consumption of the three approaches when LSTM model retraining is not required by these approaches. It is clear that RePAD2 has a similar performance as RePAD, but a slightly better performance than ReRe.

Table 4: The LSTM training ratio of different approaches on CC2-10.

| Approach | LSTM model retraining ratio |
|---|---|
| RePAD2 ($W$=1440) | 0.014 (=555/40320) |
| RePAD2 ($W$=4032) | 0.012 (=460/40320) |
| RePAD2 ($W$=8064) | 0.011 (=448/40320) |
| RePAD2 ($W$=16128) | 0.011 (=428/40320) |
| RePAD | 0.010 (=417/40320) |
| ReRe | 0.010 (=417/40320) for detector 1 |
| | 0.038 (=1522/40320) for detector 2 |

Table 5: Time consumption of different approaches on CC2-10 when LSTM model retraining is required.

| Approach | Average time to decide if a data point is anomalous (sec) | Standard deviation (sec) |
|---|---|---|
| RePAD2 (W=1440) | 0.205 | 0.030 |
| RePAD2 (W=4032) | 0.204 | 0.031 |
| RePAD2 (W=8064) | 0.200 | 0.027 |
| RePAD2 (W=16128) | 0.200 | 0.026 |
| RePAD | 0.202 | 0.030 |
| ReRe | 0.231 | 0.394 |

Table 6: Time consumption of different approaches on CC2-10 when LSTM model retraining NOT is required.

| Approach | Average time to decide if a data point is anomalous (sec) | Standard deviation (sec) |
|---|---|---|
| RePAD2 (W=1440) | 0.029 | 0.032 |
| RePAD2 (W=4032) | 0.028 | 0.012 |
| RePAD2 (W=8064) | 0.028 | 0.012 |
| RePAD2 (W=16128) | 0.028 | 0.011 |
| RePAD | 0.028 | 0.012 |
| ReRe | 0.031 | 0.080 |

## 5.2 Experiment 2

In experiment 2, we evaluated the performance of the three approaches on B3B-10. As listed in Table 7, RePAD2 had a poor precision when $W$=1440. This is because the threshold repeatedly dropped and when it dropped, its value was lower than 0.1 (see Figure 9), which caused many false positives. In other words, RePAD2 kept forgetting what it had learned before.

When $W$ was further increased to 4032 and 8064, the precision of RePAD2 significantly increased to 0.920 and 0.921, respectively. We can see from Figures 10 and 11 that the two threshold curves are more flatter than that in Figure 9, and that the values of the thresholds are higher than 0.1 in most of the time. Hence, not so many normal data points were wrongly identified as anomalous by RePAD2.

Table 7: The detection accuracy of different approaches on B3B10.

| Approach | Precision | Recall | F-score |
|---|---|---|---|
| RePAD2 (W=1440) | 0.667 | 1 | 0.800 |
| RePAD2 (W=4032) | 0.920 | 1 | 0.958 |
| RePAD2 (W=8064) | 0.921 | 1 | 0.959 |
| RePAD2 (W=16128) | 0.939 | 1 | 0.969 |
| RePAD | 0.939 | 1 | 0.969 |
| ReRe | 0.939 | 1 | 0.969 |

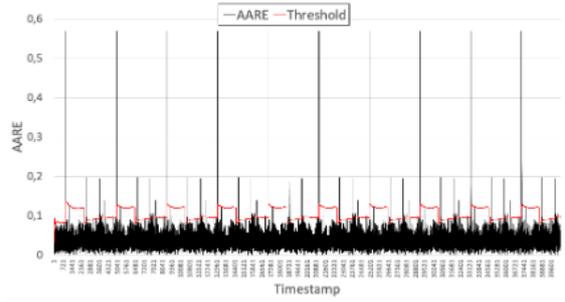

Figure 9: All derived AARE values vs. the detection threshold over time when RePAD2 worked on B3B-10 and had a sliding window of 1440.

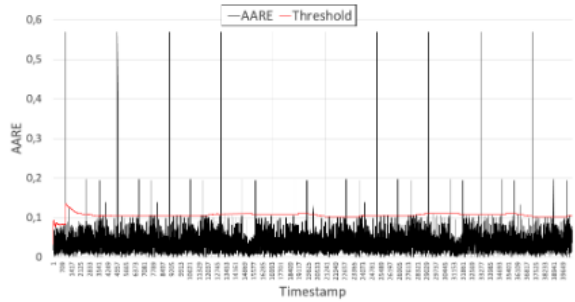

Figure 10: All derived AARE values vs. the detection threshold over time when RePAD2 worked on B3B-10 and had a sliding window of 4032.

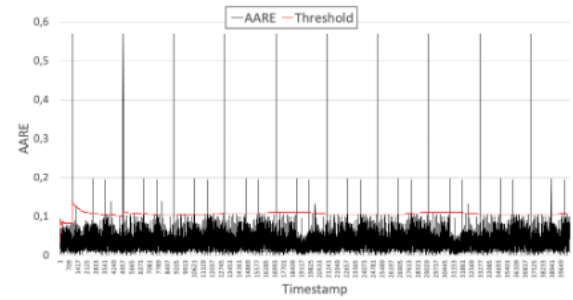

Figure 11: All derived AARE values vs. the detection threshold over time when RePAD2 worked on B3B-10 and had a sliding window of 8064.

When $W$ was further increased to 16128, RePAD2 achieved the same precision and recall (and of course the same F-score) as RePAD and ReRe. Clearly, we can see that increasing $W$ helps increase the precision of RePAD2.

Table 8 shows the LSTM retraining ratios required by the three approaches on B3B-10. Apparently, RePAD2 requires the most LSTM model retraining when $W$ is 1440, and we can also see from Table 9 that these retraining slightly impact the time consumption of RePAD2. However, when $W$ was increased, the retraining ratio of RePAD2 reduced and stabilized, and it is comparable to that of RePAD and less than that of ReRe. In addition, we can also see that the time consumption of RePAD2 (as shown in Table 9) slightly reduced as $W$ was increased, and RePAD2 is slightly more efficient than RePAD and ReRe. Table 10 shows RePAD2 had almost the same time consumption as RePAD when LSTM model retraining was not required, regardless of the value of $W$.

Table 8: The LSTM training ratio of different approaches on B3B-10.

| Approach | LSTM model retraining ratio |
|---|---|
| RePAD2 ($W$=1440) | 0.006 (=225/40320) |
| RePAD2 ($W$=4032) | 0.004 (=153/40320) |
| RePAD2 ($W$=8064) | 0.004 (=152/40320) |
| RePAD2 ($W$=16128) | 0.004 (=148/40320) |
| RePAD | 0.004 (=148/40320) |
| ReRe | 0.004 (=148/40320) for detector 1 |
| | 0.008 (=323/40320) for detector 2 |

Based on the above results on B3B-10, we conclude that RePAD2 can achieve the same detection accuracy as RePAD and ReRe when it uses the sliding window size of 16128, but it consumes less time consumption than RePAD and ReRe, especially when LSTM model retraining is required.

Table 9: The consumption of different approaches on B3B-10 while LSTM model retraining is required.

| Approach | Average time to decide if a data point is anomalous (sec) | Standard deviation (sec) |
|---|---|---|
| RePAD2 ($W$=1440) | 0.207 | 0.023 |
| RePAD2 ($W$=4032) | 0.204 | 0.019 |
| RePAD2 ($W$=8064) | 0.203 | 0.023 |
| RePAD2 ($W$=16128) | 0.203 | 0.024 |
| RePAD | 0.206 | 0.026 |
| ReRe | 0.314 | 0.706 |

Table 10: The consumption of different approaches on B3B-10 while LSTM model retraining is NOT required.

| Approach | Average time to decide if a data point is anomalous (sec) | Standard deviation (sec) |
|---|---|---|
| RePAD2 ($W$=1440) | 0.028 | 0.010 |
| RePAD2 ($W$=4032) | 0.028 | 0.009 |
| RePAD2 ($W$=8064) | 0.028 | 0.009 |
| RePAD2 ($W$=16128) | 0.028 | 0.009 |
| RePAD | 0.028 | 0.010 |
| ReRe | 0.032 | 0.133 |

# 6 CONCLUSION AND FUTURE WORK

In this paper, we have introduced RePAD2 for addressing the resource exhaustion problem that several state-of-the-art real-time and lightweight anomaly detection approaches might suffer when they work on open-ended time series for a long time. By limiting the number of historical AARE values to calculate the detection threshold that is dynamically updated at every data point (except for the first few data points), RePAD2 successfully avoids the underlying system resources from exhaustion.

Two experiments based on real-world time series from the Numenta Anomaly Benchmark have been conducted to compare RePAD2 and two other real-time and lightweight anomaly detection approaches (i.e., RePAD and ReRe). Four different sliding window sizes were used to evaluate the performance of RePAD2. According to the results, it is not recommended that RePAD2 uses a small sliding window size (i.e., using a few number of historical AARE values to calculate the detection threshold) because the detection threshold will fluctuate over time and it will cause unwanted false positives.

A large sliding window size is recommended for RePAD2. As compared with RePAD and ReRe, RePAD2 with a large sliding window can reduce false positives and increase F-score, which therefore offers either slightly better or comparable detection accuracy. In addition, RePAD2 provides a slightly better performance when it comes to the time consumption for determining whether each data point in the target time series is anomalous or not.

As our future work, we would like to implement and deploy RePAD2 on a tiny computer such as Raspberry Pi for different IoT time series anomaly detection (e.g., energy consumption, network traffic, room temperature, humidity, etc.). We also plan to deploy RePAD2 on android-based smart phones to see how it can help individual user to better monitor their network usages. In addition, we plan to extend RePAD2 to detect anomalies in multivariate open-ended time series in a real-time and lightweight manner.

# ACKNOWLEDGEMENTS

The authors want to thank the anonymous reviewers for their reviews and valuable suggestions to this paper. This work was supported by the S3UNIP project - University Grant Package Smart Software Systems funded by Høgskulen på Vestlandet (HVL) under project number 5700036-19.

# REFERENCES

Moore, R., Lopes, J. (1999). Paper templates. In *TEMPLATE'06, 1st International Conference on Template Production*. SCITEPRESS.

Smith, J. (1998). *The book*, The publishing company. London, 2nd edition.

Aggarwal, C. C. and Yu, P. S. (2008). Outlier detection with uncertain data. *In Proceedings of the 2008 SIAM International Conference on Data Mining*, pages 483–493. SIAM.

Ahmed, M., Mahmood, A. N., and Hu, J. (2016). A survey of network anomaly detection techniques. *Journal of Network and Computer Applications*, 60:19–31.

Bontemps, L., Cao, V. L., McDermott, J., and Le-Khac, N.-A. (2016). Collective anomaly detection based on long short-term memory recurrent neural networks. In *International conference on future data and security engineering*, pages 141–152. Springer.

Deeplearning4j (2023). Introduction to core deeplearning4j concepts. https://deeplearning4j.konduit.ai/. [Online; accessed 24-February-2023].

EarlyStopping (2023). What is early stopping? https://deeplearning4j.konduit.ai/. [Online; accessed 24-February-2023].

Fisher, W. D., Camp, T. K., and Krzhizhanovskaya, V. V. (2016). Crack detection in earth dam and levee passive seismic data using support vector machines. *Procedia Computer Science*, 80:577–586.

Hawkins, D. M. (1980). *Identification of outliers*, volume 11. Springer.

Hochenbaum, J., Vallis, O. S., and Kejariwal, A. (2017). Automatic anomaly detection in the cloud via statistical learning. *arXiv preprint arXiv:1704.07706*.

Hochreiter, S. and Schmidhuber, J. (1997). Long short-term memory. *Neural computation*, 9(8):1735–1780.

Laptev, N., Amizadeh, S., and Flint, I. (2015). Generic and scalable framework for automated time-series anomaly detection. In *Proceedings of the 21th ACM SIGKDD international conference on knowledge discovery and data mining*, pages 1939–1947.

Lavin, A. and Ahmad, S. (2015). Evaluating real-time anomaly detection algorithms–the numenta anomaly benchmark. In *2015 IEEE 14th international conference on machine learning and applications (ICMLA)*, pages 38–44. IEEE.

Lee, M.-C., Lin, J.-C., and Gan, E. G. (2020a). {ReRe}: A lightweight real-time ready-to-go anomaly detection approach for time series. In *2020 IEEE 44th Annual Computers, Software, and Applications Conference (COMPSAC)*, pages 322–327. IEEE. arXiv preprint arXiv:2004.02319. The updated version of the ReRe algorithm from arXiv was used in this RePAD2 paper.

Lee, M.-C., Lin, J.-C., and Gran, E. G. (2020b). {RePAD}: real-time proactive anomaly detection for time series. In *Advanced Information Networking and Applications: Proceedings of the 34th International Conference on Advanced Information Networking and Applications (AINA-2020)*, pages 1291–1302. Springer. arXiv preprint arXiv:2001.08922. The updated version of the RePAD algorithm from arXiv was used in this RePAD2 paper.

Lee, M.-C., Lin, J.-C., and Gran, E. G. (2021a). How far should we look back to achieve effective real-time time-series anomaly detection? In *Advanced Information Networking and Applications: Proceedings of the 35th International Conference on Advanced Information Networking and Applications (AINA-2021), Volume 1*, pages 136–148. Springer. arXiv preprint arXiv:2102.06560.

Lee, M.-C., Lin, J.-C., and Gran, E. G. (2021b). {SALAD}: Self-adaptive lightweight anomaly detection for real-time recurrent time series. In *2021 IEEE 45th Annual Computers, Software, and Applications Conference (COMPSAC)*, pages 344–349. IEEE.

Lee, T. J., Gottschlich, J., Tatbul, N., Metcalf, E., and Zdonik, S. (2018). Greenhouse: A zero-positive machine learning system for time-series anomaly detection. *arXiv preprint arXiv:1801.03168*.

LinkedIn (2018). linkedin/luminol [online code repository]. https://github.com/linkedin/luminol. [Online; accessed 24-February-2023].

NAB (2015). numenta/nab [online code repository]. url={https://github.com/numenta/NAB}. [Online; accessed 24-February-2023].

Ren, H., Xu, B., Wang, Y., Yi, C., Huang, C., Kou, X., Xing, T., Yang, M., Tong, J., and Zhang, Q. (2019). Time-series anomaly detection service at microsoft. In *Proceedings of the 25th ACM SIGKDD international conference on knowledge discovery & data mining*, pages 3009–3017.

Schneider, J., Wenig, P., and Papenbrock, T. (2021). Distributed detection of sequential anomalies in univariate time series. *The VLDB Journal*, 30(4):579–602.

Siffer, A., Fouque, P.-A., Termier, A., and Largouet, C. (2017). Anomaly detection in streams with extreme value theory. In *Proceedings of the 23rd ACM SIGKDD International Conference on Knowledge Discovery and Data Mining*, pages 1067–1075.

Staudemeyer, R. C. (2015). Applying long short-term memory recurrent neural networks to intrusion detection. *South African Computer Journal*, 56(1):136–154.

Twitter (2015). Twitter/anomalydetection [online code repository]. {https://github.com/twitter/AnomalyDetection}. [Online; accessed 24-February-2023].

Wu, J., Zeng, W., and Yan, F. (2018). Hierarchical temporal memory method for time-series-based anomaly detection. *Neurocomputing*, 273:535–546.

Xu, J. and Shelton, C. R. (2010). Intrusion detection using continuous time bayesian networks. *Journal of Artificial Intelligence Research*, 39:745–774.